\documentclass[letterpaper, 10 pt, conference]{ieeeconf} 

\IEEEoverridecommandlockouts % Don't forget this command!

\usepackage[noadjust]{cite} % group references, 
\usepackage{url}

\usepackage[breaklinks]{hyperref}
\usepackage{graphicx}
\usepackage{caption}
\usepackage{siunitx}
\let\labelindent\relax
\usepackage[inline]{enumitem}
\usepackage{amsmath}

\title{When Being Soft Makes You Tough: A Collision-Resilient Quadcopter Inspired by Arthropods' Exoskeletons} 

\author{
		Ricardo de Azambuja$^{1}$,
        Hassan Fouad$^{1}$,
		Yann Bouteiller$^{1}$,
		Charles Sol$^{1}$,
        Giovanni Beltrame$^{1}$

	\thanks{$^{1}$MISTLab, \'Ecole Polytechnique de Montr\'eal, Montr\'eal, Canada}
	\thanks{{\tt Contact: ricardo.azambuja@gmail.com.}}
}

\begin{document} 

\maketitle
\thispagestyle{empty}
\pagestyle{empty}

\begin{abstract}
  Flying robots are usually rather delicate and require protective
  enclosures when facing the risk of collision, while high complexity
  and reduced payload are recurrent problems with collision-resilient
  flying robots. Inspired by arthropods' exoskeletons, we design a
  simple, open source, easily manufactured, semi-rigid structure with
  soft joints that can withstand high-velocity impacts. With an
  exoskeleton, the protective shell becomes part of the main robot
  structure, thereby minimizing its loss in payload capacity. Our
  design is simple to build and customize using cheap components
  (e.g. bamboo skewers) and consumer-grade 3D printers. The result is
  CogniFly, a sub-\SI{250}{\gram} autonomous quadcopter that survives
  multiple collisions at speeds up to \SI{7}{\meter\per\second}. In
  addition to its collision-resilience, CogniFly carries sensors that allow it to fly for
  approx. \SI{17}{\min} without the need of GPS or an external motion
  capture system, and it has enough computing power to run deep neural
  network models on-board. This structure becomes an ideal platform
  for high-risk activities, such as flying in a cluttered environment
  or reinforcement learning training, by dramatically reducing the
  risks of damaging its own hardware or the environment. Source code,
  3D files, instructions and videos are available (open source license) through the
  project's website: \hyperlink{https://thecognifly.github.io/}{https://thecognifly.github.io}.
\end{abstract}

\section{Introduction}
The world is an unforgiving place and any robot will sooner or later
face a collision. Complex sensors and computational methods are
usually employed to avoid collisions, while nature takes a different
approach and, in many cases, animals embrace collisions instead of
avoiding them. One example of such amazing behaviour comes from a well
known arthropod: the cockroach. This undervalued insect is capable of
achieving faster direction transitions by hitting its head against
walls~\cite{jayaram2018transition}. Arthropods' collision-resilience
results from the presence of jointed appendages, body segmentation and a nonliving external
skeleton, called an \emph{exoskeleton}. An exoskeleton has a
dual purpose: \emph{it works as support and protective structure}. 
Nevertheless, it is not necessarily fully rigid,
mixing stiff segments connected by soft joints~\cite{Clark2018-id}.

\setlength{\belowcaptionskip}{-15pt}

\begin{figure}[t]
	\centering
	\includegraphics[width=0.85\linewidth]{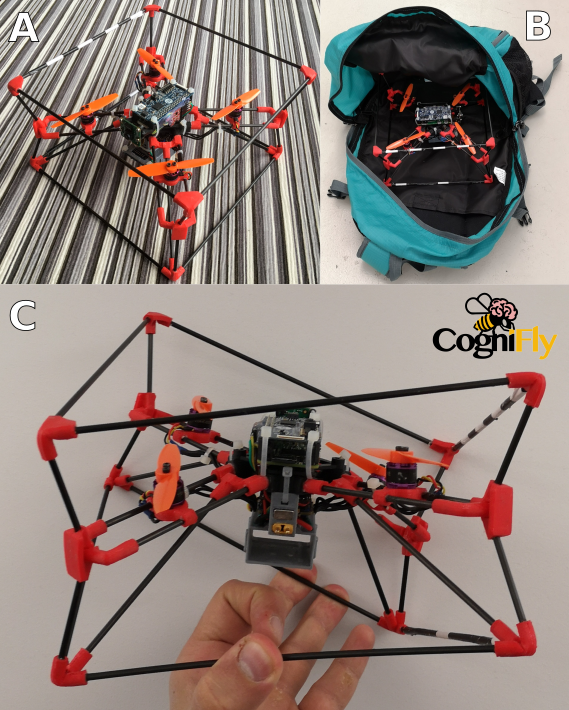}
	\caption{CogniFly (A, B and C) is a small, under-\SI{250}{\gram}, 
	      open source collision-resilient quadcopter. Its frame mixes soft (red)
		  and rigid (black/gray) parts, allowing it to better absorb and distribute impact energy.}
	\label{fig:cognifly}
\end{figure}

Uncrewed~\cite{garber2012style, koren_2019} Aerial Vehicles (UAVs) can
take advantage of collisions, too. This idea was shown to reduce the control complexity when
flying surrounded by trees~\cite{briod2014collision},
only using the sense of
touch~\cite{briod2013contact}, and to go through confined spaces by crashing onto each other
and the environment~\cite{mulgaonkar2017robust}. Recent studies have
presented contact–based navigation~\cite{khedekar2019contact} and even
a complete collision inertial odometry algorithm that uses collisions~\cite{lew2019contact}.  Collision
tolerance also was proved useful for reinforcement
learning using real robots~\cite{gandhi2017learning}.

Looking at solutions provided by nature, arthropods
can be a rich source of inspiration for innovative UAV designs. A structure inspired by their exoskeletons could replace a traditional cage by mixing rigidity with flexibility to absorb
collision energy and protect sensitive components. For a UAV, this would increase its maximum payload (useful weight it can carry), since frame and cage are now fused, and it could allow to more easily physically interact
with the environment. 

In this paper, we present the CogniFly (Fig.~\ref{fig:cognifly}): a small size, sub-\SI{250}{\gram} 
and collision resilient quadcopter. Inspired by arthropods' exoskeletons, 
it uses a semi-rigid structure with soft joints fusing frame and protective 
cage, thus providing protection against collisions and helping to maintain 
the structural integrity of the quadcopter. The CogniFly is designed with 
indoors, outdoors and subterranean exploration in mind.
The main requirements for our quadcopter design were: 
\begin{enumerate*} [label=\textbf{\roman*})]
\item Small form factor and weight (sub-\SI{250}{\gram}), for enhanced ability 
of exploring relatively narrow spaces, and easier handling and logistics
\item Enough computational power to carry out on-board image 
processing from its own camera using deep neural network models. 
\item Open source design focused on hobby grade 3D printers, 
and a software base that is easy to interact with.
\item Easy access to the battery to pave the 
way for automated battery swapping.\footnote{The battery swapping system is 
the subject of currently ongoing research.}.
\end{enumerate*}

\begin{figure}[tb]
	\centering
		\includegraphics[width=0.80\linewidth]{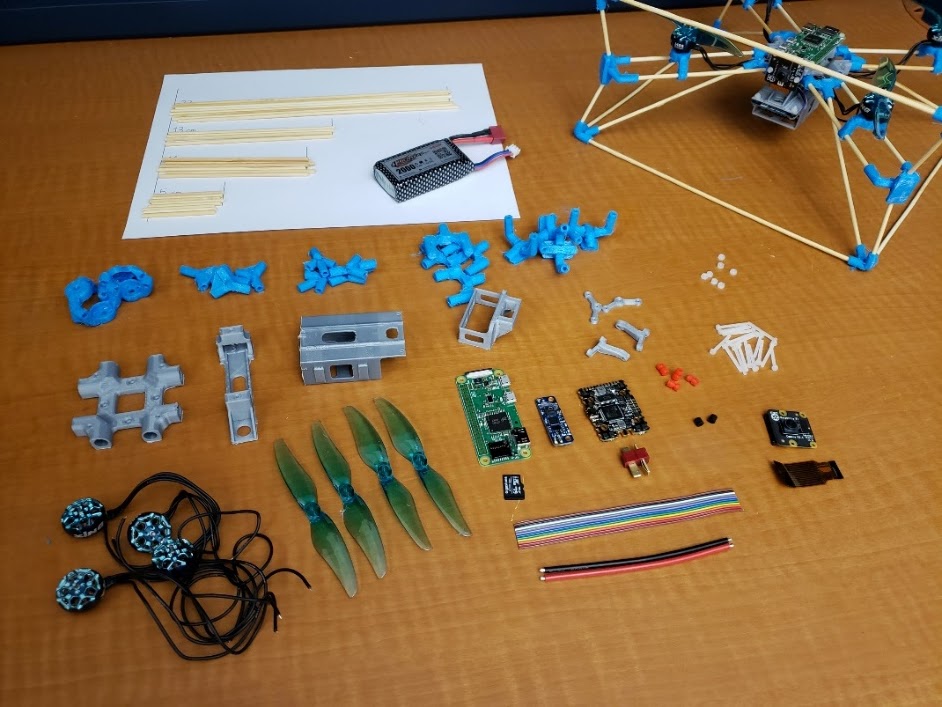}
		\caption{Bamboo version (top right), using lighter parts optimized for flight time (up to~\SI{17}{\min} 
		using 1404/3000KV motors, 4'' propellers and 2S 2000mAh LiPo battery).}
	\label{fig:cognifly_bamboo}
	\end{figure}

As an open source project, we consider having a 
customizable and easy to fabricate or repair design is paramount. 
Thus, it relies on readily available carbon fibre rods, or even bamboo skewers (Fig.~\ref{fig:cognifly_bamboo}) for an even cheaper implementation\footnote{All crash test data and models are based on the carbon fibre rod version.},
and 3D printed parts that are simple, small and easy to print.

CogniFly's exoskeleton provides protection 
for sensitive components, structural integrity for the quadcopter, 
and effective passive damping of impacts. To test its collision 
resilience, we run crash-landing experiments and compare to a rigid counterpart. 
Using these data, we model our exoskeleton 
as linear viscoelastic components (Kelvin-Voigt model~\cite{jayaram2018transition}) resulting in a
lumped mass-spring-damper model that allows us to study the
collision viability of the drone, and the role of different materials and configurations in the future.

Finally, we provide a Python library, as well as Buzz~\cite{buzz2016} 
language extension, for easy programming. 
The latest version of CogniFly is able to fly autonomously for up to \SI{17}{\min}, and run 
algorithms, such as deep neural network object detectors,
despite of its small size and sub-\SI{250}{\gram} weight. 
The CogniFly opens the doors for
potential applications like agriculture, subterranean exploration, drone swarming 
and many others.

\section{Related Work}
In general, the main strategy to endow UAV designs with collision
resilience has been the simple addition of external protective structures
like cages and bumpers
(e.g.~\cite{agha2020shapeshifter,khedekar2019contact,
  briod2014collision, kalantari2013design}). These structures
evolved into designs that allowed some level of movement to avoid external disturbances
like a sphere containing a gimbal or a cylinder capable of
rolling around its main axis (e.g.~\cite{kalantari2013design,
  briod2014collision, salaan2019development, lew2019contact, mintchev2018soft}), but those 
design choices have some drawbacks like increased weight, mechanical complexity, and 
a general lack
of energy absorption for force components perpendicular (rigid cages) or aligned (rigid and soft cages) to
the axis of rotation as the internal structures are connected to the cage using rigid parts.

Researchers have been trying to improve collision resilience for UAVs
using different strategies than traditional rigid cages.  Carbon fibre
structures are usually popular with drone frame designers because of
their steel-like stiffness. However, it is also possible to take
advantage of their elastic behaviour (Euler springs~\cite{klaptocz2013euler}) to design
flexible protective cages (e.g. ~\cite{briod2012airburr, klaptocz2013euler, 
briod2013contact}). In fact,
many cage designs that don't even claim impact energy absorption share the same elastic
behaviour, to a certain extent, as they are made of long, curved carbon fibre parts (e.g.~\cite{briod2014collision, 
kalantari2013design, agha2020shapeshifter, salaan2018close}).
Nevertheless, the high strength of carbon fibre limits its stand-alone
energy absorption applications to very long and thin struts~\cite{mintchev2017insect}, creating
a problem when the focus is designing small-sized UAVs.

Structures protecting UAVs are usually made of rigid materials, 
but that is not vital, and 
even Expanded Polypropylene~(EPP) can be used for collision resilience~\cite{gandhi2017learning}.
A weakness of materials like EPP is the low stiffness to weight ratio that 
makes such materials too heavy for high-energy
impacts~\cite{klaptocz2013euler}.
In addition to EPP, soft protective structures for UAVs can use different materials. 
By precision laser-cutting and folding 
very thin plastic sheets it is possible to build simple bumpers \cite{minicore2020} 
or an origami rotatory bumper for impacts up to 2m/s~\cite{sareh2018rotorigami}.

The weight of a UAV can vary
from a few grams to kilograms. While a pico drone weighs as few as
\SI{25}{\gram}~\cite{mulgaonkar2017robust}, the total mass of a more
complex drone using gimbals protecting each propeller easily reaches more than \SI{2}{\kg} when the batteries
are included~\cite{salaan2019development}. Still, many regulatory
agencies take the \SI{250}{\gram} value as the limit for UAVs to be considered safe.

Weight reduction is a simple collision
resilience strategy~\cite{mulgaonkar2017robust, jayaram2018transition}, but small weight and 
size comes with disadvantages such as smaller motors, limiting 
payload and ability to counter disturbances. 
Reduced payload also restricts battery size, computational 
power and ultimately many interesting applications.

A flexible frame that is capable of absorbing energy during a collision while
protecting sensitive parts, by changing its shape or employing
non-destructive deformation, is a very interesting option for collision resilient
drones. However, previous strategies based on flexible
frames (e.g.~\cite{shu2019quadrotor, mintchev2017insect,
  mintchev2018bioinspired}) make it very difficult
for the UAV to instantly recover from a hard collision because they
all lead to an inevitable fall to the ground as they automatically
fold or disconnect the motors.

One advantage of flexible frames without guards or a cage to
keep propellers from touching obstacles (e.g.~\cite{mintchev2017insect,
mintchev2018bioinspired}) is the increase in
payload capability.
However, unprotected propellers do not allow UAVs to physically
interact with the external world, even considering the use of special
flexible propellers~\cite{jang2019design}, as the decrease in thrust
and torque from a bent propeller could easily
destabilize the UAV.

\begin{figure}[t]
	\centering
	\includegraphics[width=1.0\linewidth]{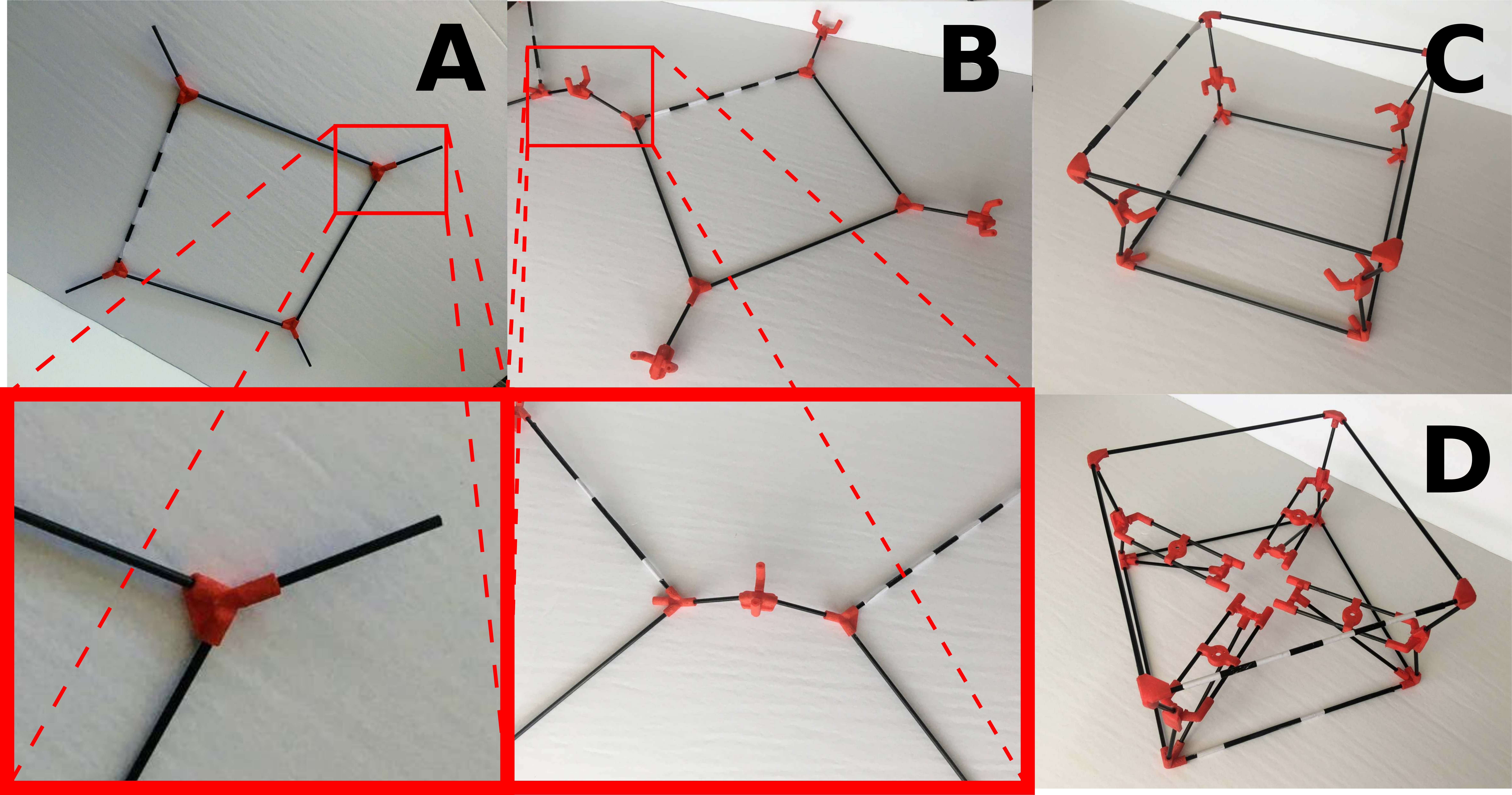}
	\caption{First, top and bottom are built~(A).  
	Corners are inserted and both sides are connected~(B) and folded~(C). 
	Finally, arms and extra dampening struts are attached~(D).}
	\label{fig:folding_as_you_assemble}
\end{figure}

Although other designs~\cite{kornatowski2017origami, mintchev2018soft, 
zha2020collision} share some characteristics also seen in the CogniFly, 
they do not employ a truly flexible exoskeleton. Some designs have external tensegrity 
structures where soft, flexible parts are used to connect more rigid 
components to form a flexible protective cage, but \cite{mintchev2018soft} 
uses a rigid rotating axis at its center, and \cite{zha2020collision} a 
rigid quadcopter x-frame, making those two UAVs only flexible to the extent 
of their cages. The cargo drone from~\cite{kornatowski2017origami}, 
according to the publicly available information, is fully flexible only 
in its folded configuration (for storage) as the quadcopter arms are 
secured together (screw system) before flight transforming the arms 
into a rigid x-frame.

Only a few previous works explored truly flexible frames, 
where the UAV arms holding the motors are not rigidly connected to 
the central part of the frame
(e.g.~\cite{mintchev2017insect, mintchev2018bioinspired, minicore2020}), but the flexible arms
mainly offer protection against \emph{in-plane} collisions while
leaving propellers mostly exposed (hindering the ability to physically
interact with the external world~\cite{mintchev2017insect, mintchev2018bioinspired}, 
or protecting propellers in only one plane~\cite{minicore2020}, and they are not easy to
manufacture without special materials or tools. Still, 
collision-resilient works under \SI{250}{\gram} show bare minimum
payload capacity, restricting their on-board computational power and
their ability to fly autonomously without an external computer or
motion capture system. Finally, their batteries are designed to be
manually connected and extracted by human hands, making the use of an
automatic battery swapping station very unlikely.

From all previous collision-resilient UAV designs, only a few manage
to keep the total weight below the \SI{250}{\gram}
threshold (e.g. \cite{mulgaonkar2017robust, sareh2018rotorigami,
  mintchev2017insect, mintchev2018bioinspired, minicore2020}), and, besides~\cite{minicore2020}, those employ
high-speed coreless brushed DC motors, limiting their payload, total
flight time and lifespan~\cite{mulgaonkar2014power}. Moreover, those
sub-\SI{250}{\gram} UAVs have a very limited maximum payload, on-board
computing (if any) and sensing capacity, requiring external control and/or an
external motion capture systems. 

\section{UAV Design}

\subsection{Structural design}

In our design, we opt for a structure, loosely inspired by arthropods' exoskeletons, mixing rigid and soft components. 
The fragile on-board electronics are mounted on rigid parts (3D printed in ABS, black or PLA, gray, Figs.~\ref{fig:cognifly} 
and \ref{fig:cognifly_bamboo}) that are placed at the central gap of the exoskeleton~(Fig.~\ref{fig:folding_as_you_assemble}-D). However, 
these rigid 3D printed parts 
are connected to the exoskeleton using flexible joints (red and blue parts, Fig.~\ref{fig:cognifly}, \ref{fig:cognifly_bamboo} 
and \ref{fig:folding_as_you_assemble}). 
Moreover, we mount the motors on special flexible parts to allow them to flex during impacts.

The exoskeleton~(Fig.~\ref{fig:folding_as_you_assemble}-D), that can be made of carbon fibre (Fig.~\ref{fig:cognifly})
or bamboo (Fig.~\ref{fig:cognifly_bamboo}), gives the quadcopter a final box-like external shape.
By having \emph{flat-like} external faces, our quadcopter can take advantage of its collision resilience
to easily align itself against external structures that are big enough compared to the gaps between the outermost rods.

To control where the parts should bend, we designed 3D printed soft joints to interconnect the 
rigid parts of the exoskeleton~(see detail in Fig.~\ref{fig:folding_as_you_assemble}-A). 
These joints use flexible filament (TPU 95A), providing sufficient damping for impacts, and helping CogniFly survive impacts at speeds 
up to \SI{7}{\meter\per\second} (carbon fibre version, Fig.~\ref{fig:cognifly}). TPU 95A enables the drone
to be generally flexible, as shown in Fig.~\ref{fig:cognifly}-C, 
while keeping the integrity of its central rigid part.  The choice of the exoskeleton dimensions 
becomes a trade-off between reducing the probability of direct impact on sensitive components and 
general total size and weight. In the case of CogniFly, we wanted to make it
small enough to fit in a standard backpack, thus it measures only \SI{210}x\SI{210}x\SI{120}{\mm}.

The final weight distribution for the carbon fibre version: 
\begin{enumerate*}[label=\textbf{\roman*})] 
	\item Exoskeleton shown in Fig.~\ref{fig:folding_as_you_assemble}-D:~\SI{62}{\g}.
	\item Central ABS parts:~\SI{25}{\g}. 
	\item Quadcopter without battery (Fig~\ref{fig:cognifly}-C):~\SI{178}{\g}. 
	\item Battery:~\SI{58}{\g}.
\end{enumerate*}
Total weight (\SI{178}{\g}+\SI{58}{\g}): \SI{236}{\g}.

\begin{figure}[t]
	\centering
	  \includegraphics[width=0.9\linewidth]{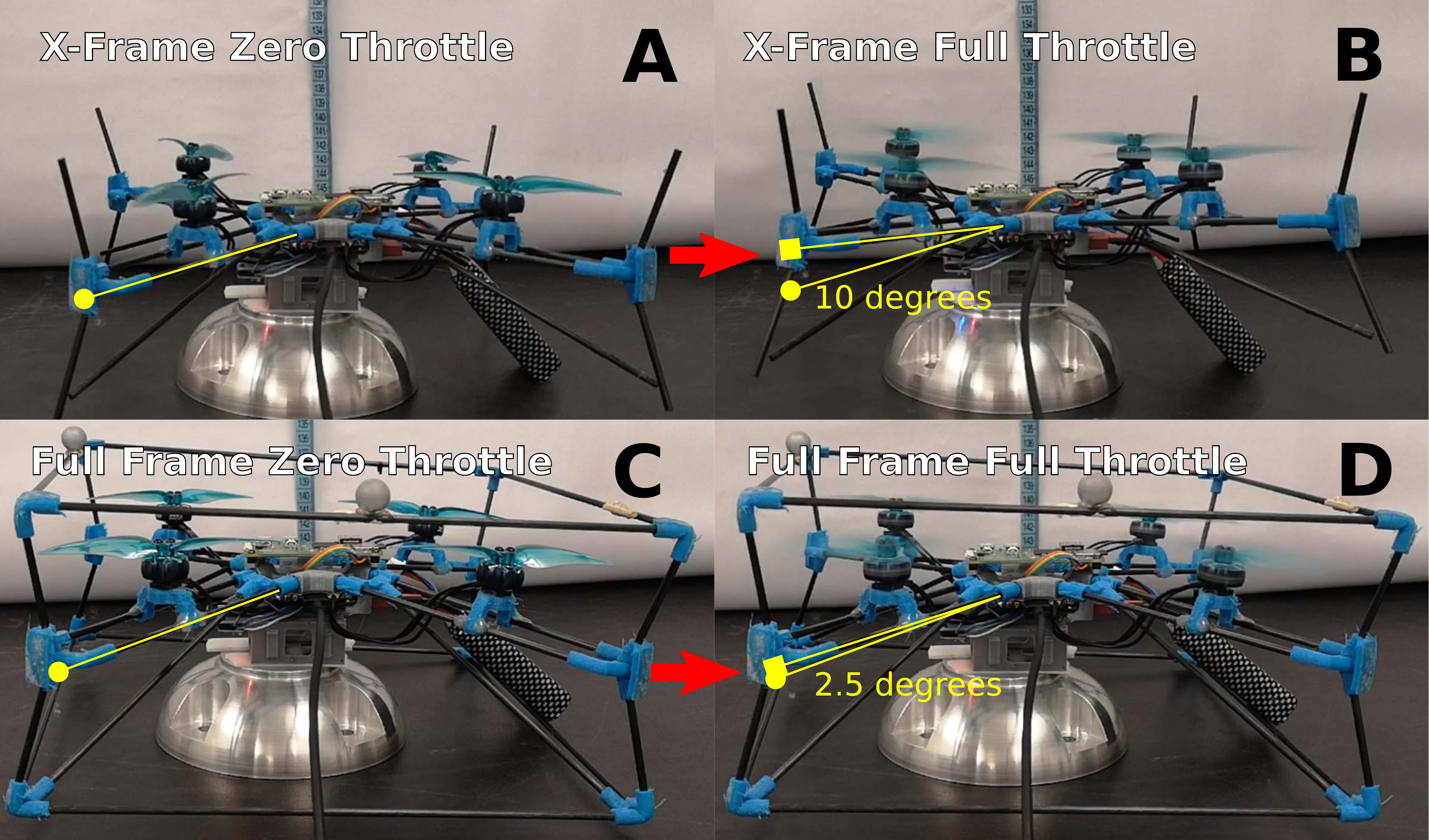}
	%   \captionsetup{belowskip=-20pt}
	  \caption{Comparison between the equivalent \emph{x-frame} (A/B) and the \emph{full frame} (C/D).
	  The \emph{x-frame} (B) bends 4x more than the \emph{full frame} (D) for the same throttle value.}
	\label{fig:cognifly_exoskeleton_flexing}
\end{figure}

To visualize the importance of the exoskeleton for load distribution, 
Fig.~\ref{fig:cognifly_exoskeleton_flexing} shows two configurations under full thrust and rigidly attached to the desk only by the battery holder: with (\emph{full frame}, Fig.~\ref{fig:cognifly_exoskeleton_flexing}-C and D) 
and without (\emph{x-frame}, Fig.~\ref{fig:cognifly_exoskeleton_flexing}-A and B) the
external protective parts of the exoskeleton.

\subsection{Manufacturability}

The main aspects for assessing the
manufacturability that we adopt are:
\begin{enumerate*}[label=\textbf{\roman*})]
\item Accessibility to different structural components.
\item Required manufacturing processes and facilities.
\item Cost.
\end{enumerate*}
The main components of the exoskeleton are carbon
fibre or bamboo rods and joints made of TPU 95A~(Fig.~\ref{fig:folding_as_you_assemble}).

Carbon fibre rods are cheap, readily available, easy to cut
and modify, and there are several examples of previous
works using carbon fibre rods in UAVs (e.g. ~\cite{briod2014collision,
  klaptocz2013euler, zha2020collision}).  On top of that, our design
allows the use of bamboo rods (BBQ
skewers, Fig.~\ref{fig:cognifly_bamboo}), at the expense of lower impact
resistance.

A low-cost desktop 3D printer
(Monoprice Mini v2)
was used for all parts, hence we were constrained to small and simple parts, and 
all flexible parts are printed flat and without supports. 
Moreover, we took advantage of the TPU95A flexibility and designed the parts to
work as living hinges and
bend~(Fig.~\ref{fig:folding_as_you_assemble}-A) or snap-fit as the drone
is assembled.

\subsection{On-board hardware and software tools}
CogniFly uses as its main
controller a single-board computer running Linux. Additionally, a
cheap and popular flight controller (e.g. Kakute F7 Mini) running our customized version of
iNAV takes care of lower level flight control tasks. 

For running deep neural models on-board,
CogniFly offers three possible
configurations: two using the Raspberry Pi Zero~W as
its high level controller together with Google AIY Vision or Coral USB; 
a third option uses only a Google Coral Dev Mini board.

In addition to a normal camera, CogniFly carries Optical Flow (PMW3901) and
Time-of-Flight (VL53L0X) sensors, thus allowing it to navigate
autonomously.

Finally, we developed open source tools to allow CogniFly
to operate autonomously and run
complex DCNN models (YAMSPy), to be remote controlled from an external
computer (cognifly-python), mocap extension for
iNAV, and swarm experiments using Buzz~\cite{buzz2016} (cognifly-buzz).

\section{Collision Resilience Experiments}

\begin{figure}[tb]
	\centering
	\includegraphics[width=0.85\linewidth]{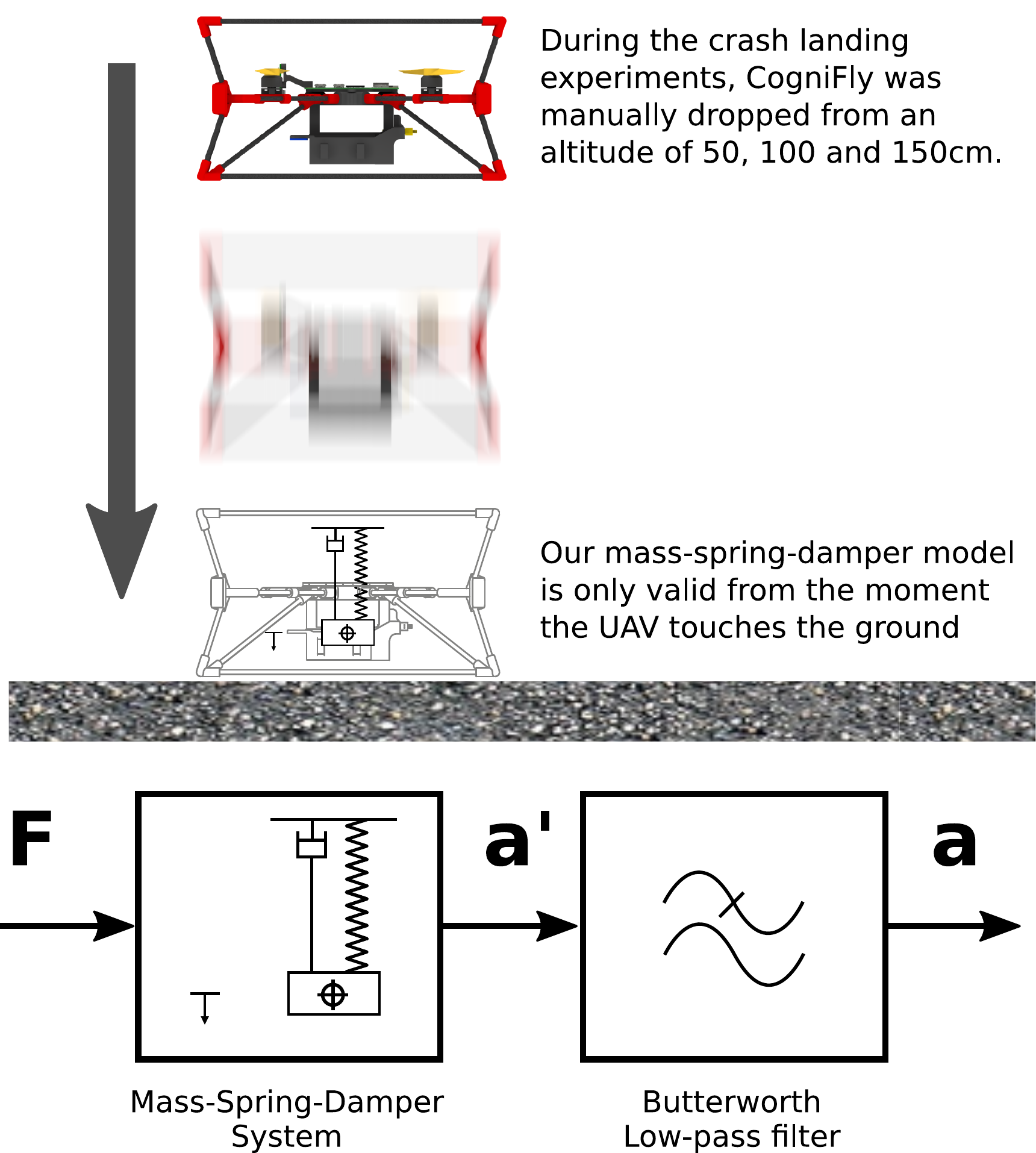}
	% \captionsetup{belowskip=-20pt}
	\caption{The model is valid from the moment the exoskeleton touches the floor
		($x_0=0mm$ and $v_0=\sqrt{2*g*h}$, where $h$ is the drop
		altitude) until $x=16mm$ and the battery collides.}
	\label{fig:experiment}
\end{figure}

We perform a series of crash-landing (free fall) tests~(Fig.~\ref{fig:experiment}), using the carbon fibre version of our design~(Fig.~\ref{fig:cognifly}), 
and record the results as 
absolute acceleration (Eq.~\ref{eqn:abs_acc}) to avoid the need of a complex 
guiding/alignment system that would be otherwise necessary to isolate individual axis during the experiments. 
These tests highlight the ability
of our flexible exoskeleton to absorb impact energy by comparing the
acceleration recorded from a CogniFly made with carbon fibre rods (Fig.\ref{fig:cognifly},
\SI{241}{\gram} when fitted with the datalogger) and only the rigid central part of the frame, which
holds the flight controller, single-board computer, sensors and the battery, made entirely of ABS~(a
bag of loose screws was added to reach \SI{239}{\gram}).

\begin{equation}
	|acc| = \sqrt{{acc_x}^2+{acc_y}^2+{acc_z}^2}
	\label{eqn:abs_acc}
\end{equation}

We reckon a vertical free fall to be a critical scenario as we
consider payload contact with hard exterior objects, like the ground,
has the highest potential of causing damage because the absolute
acceleration (Eq.~\ref{eqn:abs_acc}) peaks in such cases.  Moreover, the battery
is located at the bottom part of the drone, and it should not be subjected to
extreme loads.

Acceleration values saved by flight controllers are usually limited to
$\pm16G$ with heavy filtering smoothing any peaks, therefore, we use a
custom datalogger for our experiments. Our datalogger setup has an accelerometer (ADXL377, 3-Axis, $\pm200G$, \SI{500}{\Hz}) 
rigidly attached to the center of the frame, and a Feather M0 Express running a custom firmware 
for sampling (1kHz) the sensor and saving the data.

\subsection{Mass-spring-damper model}
We model the impact absorbing aspect of the CogniFly
as linear viscoelastic components (Kelvin-Voigt model~\cite{jayaram2018transition}) 
resulting in a
lumped mass-spring-damper system that is described by
\begin{equation}
m\ddot{x} + c\dot{x} + kx = F
\label{eqn:msd}
\end{equation}
where $m>0$ is the mass of the drone, and $c>0$ and $k>0$ are the equivalent
damping and stiffness coefficients, respectively.
Moreover, we augment the model with a first order
Butterworth low-pass filter with a cutoff frequency of \SI{500}{\Hz}
to take into account the sampling latency of our accelerometer setup.

To find the parameters, we put Eq.~\ref{eqn:msd} in the following form
\begin{equation}
\begin{bmatrix}
\dot{x}\\\dot{v}
\end{bmatrix} = \begin{bmatrix}
0&1\\-\frac{k}{m}&-\frac{c}{m}
\end{bmatrix}\begin{bmatrix}
x\\v
\end{bmatrix}+\begin{bmatrix}
0\\\frac{1}{m}
\end{bmatrix}F
\label{eqn:fos}
\end{equation} 
and then use Scipy signal processing tool \emph{lsim} to solve the
system \eqref{eqn:fos} to obtain the velocity and displacement of the
payload's centre of gravity as a function of the initial displacement ($x_0$),
velocity ($v_0$), and the parameters $k,c$ to be estimated. In order to model
the conditions at moment of impact, we set the external force $F$ to
gravity ($mg$), the initial displacement to zero and the initial
velocity to the value of velocity just before impact (without air drag).

The equivalent stiffness $k$ was obtained by deforming
the payload to a known displacement, while measuring the applied
force. From these data, we fit a linear
model constrained to the minimum force before any deformation could be
measured ($x=0mm$ and $F=mg=2.36N$).

The equivalent damping $c$ is estimated by minimizing the Mean Square 
Error between the mass-spring-damper
model~(Eq.~\ref{eqn:msd}), after passing through the low-pass filter, 
and the collected acceleration data from
the end of the free fall until the peak of the measured absolute
acceleration (Eq.~\ref{eqn:abs_acc}) for all experiments ($50$, $100$
and \SI{150}{\cm}). However, as the number of trials for each
experiment is different ($101$, $97$ and $89$, respectively), the
final value is weighted accordingly. 

Minimization was carried out using Scipy
Optimize \emph{minimize}, with \emph{Nelder-Mead} method, default
arguments, \emph{cauchy} loss ($L(z)=ln(1+z)$), and initial values of $c=50$ and $k=7040$. It resulted in the 
coefficients $c=46.32$ and
$k=6996.12$. However, as the calculated value for $k$ was very close to the
static one experimentally measured, we adopted the
coefficients $c=46$ and $k=7040$ for our model.

\begin{figure}[tb]
	\centering
	\includegraphics[width=0.9\linewidth]{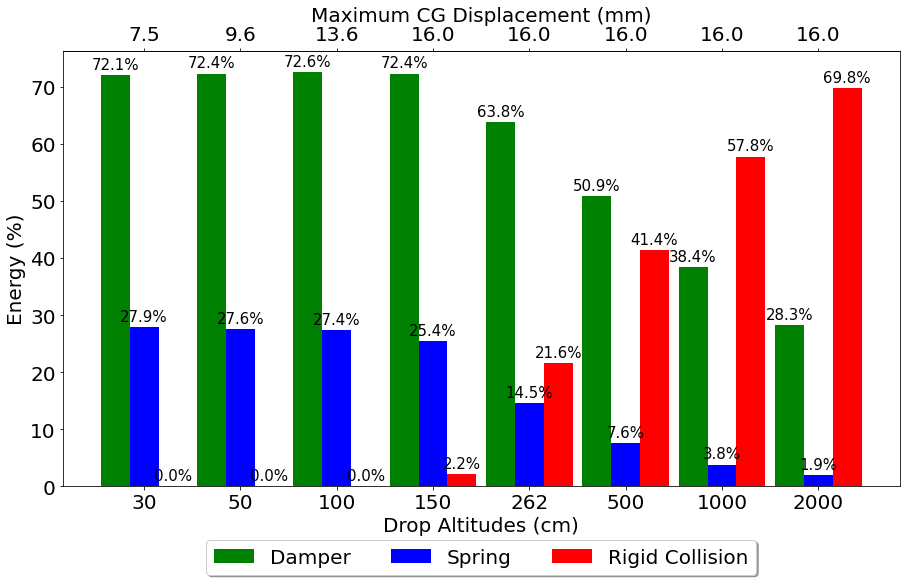}
	\caption{Energy distribution according our model.}
	\label{fig:model_energy_bars}
\end{figure}

We use the proposed model to calculate the percentages of the energy
that go into different parts of the system during the impact, which is
depicted in Fig.~\ref{fig:model_energy_bars}.  To construct such plot,
we consider the kinetic energy ($E_k=\frac{1}{2}mv^{2}$) 
at the beginning of the impact (end of the free fall) as being the
total energy of the system. Therefore, we have two possible
situations: the battery holder never touches the ground ($x<16mm$) or
the payload hits the ground ($x\geq16mm$).

When $x<16mm$ 
(drop altitudes up to 100cm), the final kinetic energy at the point of
maximum displacement is zero (the movement is about to reverse) and
the total energy is split between stored in the spring
($E_s=\frac{1}{2}kx^2$) and dissipated by the damper ($E_d=E_k-E_s$).

Our mass-spring-damper-model is not valid for direct collisions
between the battery holder and the ground (drop altitudes from
\SI{150}{\cm} and above) and it is only valid until
$x<16mm$. Therefore, in these situations we calculate the energy
dissipated by the damper considering the difference between the
initial kinetic energy ($E_k$) and the kinetic energy when
$x=16mm$. This way, we know, in the \emph{worst} scenario, the energy
that will be dissipated during the \emph{rigid collision} (battery
holder hits the ground) will be the same as the kinetic energy
available at $x=16mm$ (represented by the red bars in
Fig.\ref{fig:model_energy_bars}) to show the severity of the
impact to the ground.

\subsection{Experimental Results}

\subsubsection{Impact testing}

\begin{figure}[tb]
	\centering
	\includegraphics[width=1\linewidth]{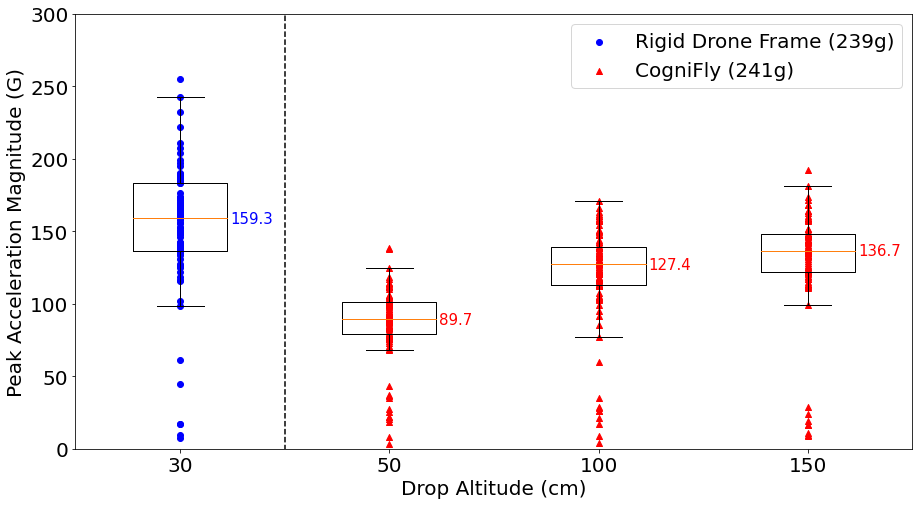}
	\caption{Experimental crash landing peak acceleration data..}
	\label{fig:crash_landing_results}
\end{figure}

We carry out the tests at three different
altitudes: \SIlist{50;100;150}{\cm}. To obtain usable data and avoid
irreparable damage to the rigid version tested, we had to limit its free fall to
\SI{30}{\cm}. Fig.~\ref{fig:crash_landing_results} shows that the
median of the absolute acceleration (Eq.~\ref{eqn:abs_acc}) peak values for the rigid frame
falling from \SI{30}{\cm} is higher than that of the CogniFly falling
from \SI{150}{\cm}. This strongly suggests our exoskeleton design is more capable of dissipating impacts than a
rigid structure made of ABS plastic with a smaller weight.

As a final experiment, we tested CogniFly (carbon fibre rods) by dropping it from the maximum
altitude our experimental setup allowed us (literally, our ceiling).
CogniFly falls from \SI{262}{\cm}, without suffering
any damage (speed at impact of approximately \SI{7}{\meter\per\second}).
Compared to some of the latest works on collision resilience UAVs with 
equivalent size and weight \cite{sareh2018rotorigami, shu2019quadrotor, 
mintchev2017insect, zha2020collision}, CogniFly reaches a higher
collision speed without suffering any damage.

\subsubsection{Maximum absolute acceleration}
One of the main uses of the exoskeleton is to provide
protection against high acceleration (deceleration) values to vulnerable components
during impacts. To simplify the necessary experimental setup, the main criterion we adopt is the
maximum absolute acceleration during a vertical
free fall impact (i.e. crash landing)~Fig.~\ref{fig:experiment}.

In addition to surviving falls, during pilot tests~(see video) our flexible exoskeleton showed the
ability to withstand in-flight frontal (vertical) collisions. Unlike~\cite{mintchev2017insect,mintchev2018bioinspired},
where the drone has to land before it is able to fly again because its
motors are disconnected from the main body during collisions, CogniFly keeps flying (e.g. bouncing off walls). 

\subsubsection{Mass-spring-damper model}
\begin{figure}[tb]
	\centering
	\includegraphics[width=0.9\linewidth]{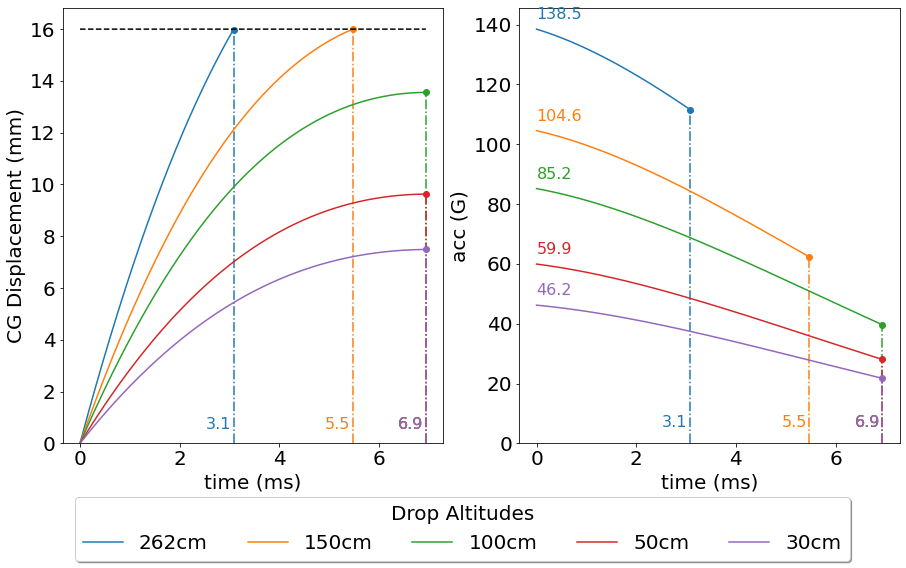}
	% \captionsetup{belowskip=-20pt}
	\caption{Predicted displacements~(left) and accelerations~(right) without
          low-pass filter and no \emph{ridig impacts}.}
	\label{fig:model_cg_displacement}
\end{figure}

We model the CogniFly with its exoskeleton structure as a
mass-spring-damper system (Fig.~\ref{fig:experiment}), with the aim of predicting the distribution
of energy stored and dissipated (Fig.~\ref{fig:model_energy_bars}),
as well as the displacement of the main payload after the beginning of
the impact until the point the acceleration reaches its maximum value
(Fig.~\ref{fig:model_cg_displacement}).

\begin{figure}[tb]
	\centering
	\includegraphics[width=0.9\linewidth]{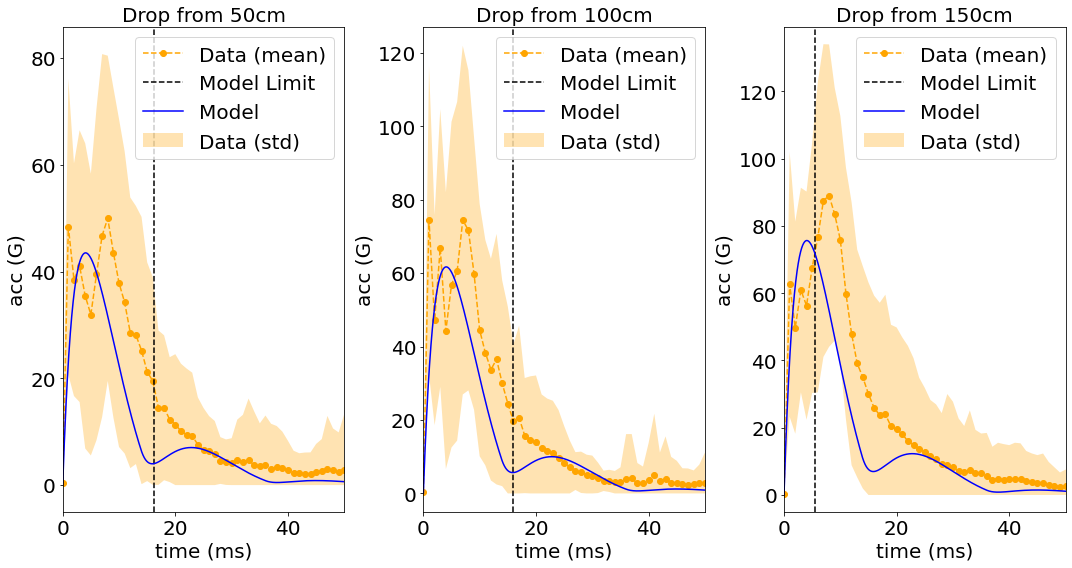}
	\captionsetup{belowskip=-0pt}
	\caption{Raw measurements vs. our model.}
	\label{fig:model_vs_data}
\end{figure}

We assess our model by comparing the accelerometer data against
the values of acceleration we predict~(Fig.~\ref{fig:model_vs_data}). 
Accelerations are used because it is challenging to devise
an affordable and reliable method for measuring the displacement of the center of
gravity during impact, while we have easy access to precise
accelerometer data. From a simple visual inspection
of Fig.~\ref{fig:model_vs_data}, the predicted values follow
the same trend as the experimental data (mean) for valid displacements~($x<16mm$, 
vertical dashed black line).

Since the main motivation behind the model is to predict the most
critical failure mode (i.e. battery holder direct hit), Fig.~\ref{fig:model_cg_displacement} shows the
predicted payload's centre of gravity displacement. The allowable
displacement for crash-landing experiments presented in this paper
(i.e. maximum distance before
the battery holder hits the ground) is \SI{16}{\mm}, and
Fig.~\ref{fig:model_cg_displacement} predicts direct impacts on the
battery holder for falls from altitudes $\geq$\SI{150}{\cm},
matching experimental results.

One of the uses for the final mass-spring-damper model is to analyse 
the energy distribution in different parts of the
drone for different altitudes, with the ability to, even if roughly, predict such
distribution for higher altitudes~(Fig.~\ref{fig:model_energy_bars}).
We show the kinetic energy at impact is distributed in
different components: stored in the spring (in blue), dissipated by the damper (in green), 
and the remaining energy that goes into \emph{rigid collision} (in red) for higher altitudes when the
payload displacement is beyond the safe allowable value. 

For altitudes below \SI{150}{\cm}, Fig.~\ref{fig:model_energy_bars}
shows that the total kinetic energy is divided only between the damper and
the spring, while for higher altitudes the amount of energy that goes
into what we call here \emph{rigid collision} increases with altitude. Such collision energy
can give an indication of how strong the impact between payload and
ground is, helping to decide how far the operational altitude can be pushed.

\section{Discussion and Conclusions}
In this paper, we introduce a new collision resilient
quadcopter design inspired by the flexible exoskeleton of arthropods, fusing
the protective cage and the main frame in one semi-rigid structure
with soft joints that can withstand high-velocity impacts. Our
UAV~(CogniFly, Fig.~\ref{fig:cognifly}) weighs under \SI{250}{\gram} and blends
rigid and soft materials, giving the final structure the ability to
absorb and dissipate impact energy, while still being sufficiently
stiff to fulfill its mission. Thanks to its exoskeleton, it is possible to
save precious weight when compared to a traditional
protective cage design.

CogniFly survived multiple collisions at speeds up to
\SI{7}{\meter\per\second} while carrying enough computing power to run
deep neural network models. Throughout a series of simple
crash-landing experiments (Fig.~\ref{fig:experiment}), we show
CogniFly withstands up to a five fold increase in the maximum
collision energy when compared to a rigid system (3D printed on ABS)
of similar weight. Moreover, we employ the experimental data to create
a lumped mass-spring-damper model that allows us to extrapolate the
results to untested cases while the calculated damping and stiffness
can be used to better understand the role of different materials or
configurations. We also make available software to allow easy of use and customization.

We designed CogniFly from the ground up for easy manufacturing and it
can be built using a very small consumer-grade 3D printer, in addition
to inexpensive off-the-shelf parts. 
The design of the drone itself was restricted by maximum weight (below
\SI{250}{\gram}) and size (fits in a backpack, Fig.~\ref{fig:cognifly}-B).
Also, considering that batteries
correspond to 33\% of UAV's total mass on
average~\cite{mulgaonkar2014power}, its battery holder and lid were designed to enable
easy manipulation of batteries, which we plan as being a stepping
stone towards designing small-sized portable battery swap stations for
extended energy autonomy.

As an interesting side effect, we noticed an increased life span of
the propellers used during our experiments. Throughout a period of
around one year crashing prototypes against walls, furniture and
floors, we only used two sets of propellers (Gemfan 3025 3X2.5,
Polycarbonate) with the second set seen in Fig.\ref{fig:cognifly}.
One explanation for that is the flexibility of CogniFly's
exoskeleton. Even the motors themselves are mounted on parts 3D printed
in flexible filament, increasing the time of impact and
reducing forces, resulting in longer life spans for propellers. 

Future work possibilities
would be extending the model to take into account collisions from
other directions, study which components or buiding methods contribute 
the most for the impact energy absorption, tune the design of the soft parts to improve its
collision resilience, verify the effectiviness of the use of a flexible net,
and analyze the impact of not being strictly
stiff in the power consumption and dynamic reactions during
flight. Ultimately, fatigue probably plays an important role in the
structure's lifespan because some parts work as living
hinges. Therefore, this would be another interesting topic to be
further studied.

\section*{Acknowledgments}
We would like to thank the financial support received from
\href{https://ivado.ca}{IVADO} (postdoctoral scholarship 2019/2020)
and the productive discussions and help received from current and past
students and interns from \href{https://mistlab.ca}{MISTLab}.

% \addtolength{\textheight}{-12cm}  % This command serves to balance the column lengths
                                  % on the last page of the document manually. It shortens
                                  % the textheight of the last page by a suitable amount.
                                  % This command does not take effect until the next page
                                  % so it should come on the page before the last. Make
                                  % sure that you do not shorten the textheight too much.

\bibliography{CogniFly}
\bibliographystyle{IEEEtran}

\end{document}